\documentclass[runningheads]{llncs}
\usepackage[T1]{fontenc}
\usepackage[misc]{ifsym}
\usepackage{graphicx}
\usepackage{subcaption}
\usepackage{algorithm,algorithmic}
\usepackage{multirow}
\usepackage{makecell}
\usepackage{enumerate}
\usepackage{float}
\usepackage{bm}
\usepackage{amssymb}
\usepackage{caption}
\usepackage{setspace}
\usepackage{amsfonts}
\usepackage{amsmath}
\usepackage{booktabs}
\usepackage{tablefootnote}
\usepackage{threeparttable}
\usepackage{hyperref}
\usepackage{url}
\usepackage[misc]{ifsym}
\begin{document}
\title{Towards Few-shot Self-explaining Graph Neural Networks}

\author{Jingyu Peng\inst{1} \and Qi Liu\inst{1,2} \and Linan Yue\inst{1} \and Zaixi Zhang\inst{1} \and Kai Zhang\inst{1}\Letter \and Yunhao Sha\inst{1}}

\authorrunning{J. Peng et al.}

\institute{State Key Laboratory of Cognitive Intelligence, \\University of Science and Technology of China, Hefei, China  \\  \and 
Institute of Artificial Intelligence \\Hefei Comprehensive National Science Center, Hefei, China  \\ 
\email{\{jypeng28,lnyue,zaixi,percy\}@mail.ustc.edu.cn}
\email{\{qiliuql,kkzhang08@ustc.edu.cn\}}}

\tocauthor{Jingyu Peng, Qi Liu, Linan Yue, Zaixi Zhang, Kai Zhang, Yunhao Sha}
\toctitle{Towards Few-shot Self-explaining Graph Neural Networks}

\maketitle              

\begin{abstract}
Recent advancements in Graph Neural Networks (GNNs) have spurred an upsurge of research dedicated to enhancing the explainability of GNNs, particularly in critical domains such as medicine. A promising approach is the self-explaining method, which outputs explanations along with predictions. However, existing self-explaining models require a large amount of training data, rendering them unavailable in few-shot scenarios. To address this challenge, in this paper, we propose a \textbf{M}eta-learned \textbf{S}elf-\textbf{E}xplaining GNN (MSE-GNN), a novel framework that generates explanations to support predictions in few-shot settings. MSE-GNN adopts a two-stage self-explaining structure, consisting of an \emph{explainer} and a \emph{predictor}. Specifically, the \emph{explainer} first imitates the attention mechanism of humans to select the explanation subgraph, whereby attention is naturally paid to regions containing important characteristics. Subsequently, the \emph{predictor} mimics the decision-making process, which makes predictions based on the generated explanation. Moreover, with a novel meta-training process and a designed mechanism that exploits task information, MSE-GNN can achieve remarkable performance on new few-shot tasks. Extensive experimental results on four datasets demonstrate that MSE-GNN can achieve superior performance on prediction tasks while generating high-quality explanations compared with existing methods. The code is publicly available at \url{https://github.com/jypeng28/MSE-GNN}.

\end{abstract}

\section{Introduction}
Due to the widespread presence of graph data in diverse domains \cite{zhang2022hierarchical,zhang2024fedgt}, Graph Neural Networks (GNNs) \cite{kipf2016semi,wu2022graph,dwivedi2023benchmarking} are attracting increasing attention from the research community. Leveraging the message passing paradigm, GNNs have exhibited remarkable efficacy across multiple scenes, including molecule property prediction \cite{wieder2020compact}, social network analysis \cite{bian2020rumor,zhang2021eatn}, and recommender system \cite{chen2020revisiting}. Despite these successes, a significant drawback of GNN models is their lack of explainability, making it unavailable for humans to understand the basis of predictions. This limitation undermines the complete trust in GNN predictions, consequently restricting their application in high-stake scenarios including medical \cite{zhang2022protgnn} and finance \cite{pourhabibi2020fraud} fields. Furthermore, the European Union has explicitly emphasized the necessity of explainability for trustworthy AI in \cite{smuha2019eu} and any studies focusing on explainability have been conducted on interpretability in other fields \cite{yue2022dare,yue2023interventional}. Therefore, there is an immediate and pressing need for research into the explainability of GNNs.

\begin{figure}[t]
	\centering
    \includegraphics[width=8cm]{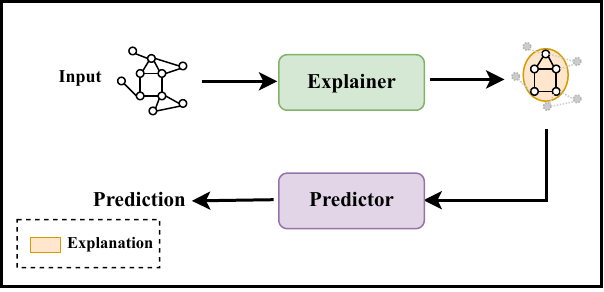}
    \caption{Paradigm of ``\emph{explainer-predictor}'' two-stage self-explaining models. The first part is composed of a \emph{explainer} which selects an explanation subgraph for each input graph. The second part is a \emph{predictor} which makes predictions based on the explanation subgraph. Given an input example ~\protect\includegraphics[width=10.pt]{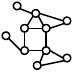} from Synthetic dataset\cite{ying2019gnnexplainer}, \emph{explainer} select  \protect\includegraphics[width=6.pt]{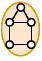} as explanation, then \emph{predictor} predicts $\hat{y}=house$ based on \protect\includegraphics[width=6pt]{fig/example_paradigm.pdf}.}
    \label{paradigm}

\end{figure}

The field of GNN explainability has witnessed substantial scholarly attention \cite{liu2022graph,muller2022dt+,sui2022causal,lin2022orphicx}. Generally, research on the explainability of GNN can be divided into two main categories: post-hoc explanations and self-explaining methods \cite{yuan2022explainability}. Among them, the post-hoc explanation strives to elucidate the predictions made by a trained GNN model. Typically, this is achieved by leveraging another explanatory model to select a subset of input as the explanation for GNN prediction. Despite their utility, these post-hoc explainers often fall short of revealing the actual reasoning process of the model \cite{rudin2018please} and require optimization for each input graph, which is time-consuming. Therefore, in this paper, we focus on self-explaining methods.

The self-explaining method refers to intrinsically explainable GNN models that offer predictions and explanations concurrently, with the prediction being rooted in the explanation. One prevalent type of self-explaining model typically adopts a ``\emph{explainer-predictor}'' two-stage paradigm, as illustrated in Figure \ref{paradigm}. This paradigm contains two stages, one is called the \emph{explainer}, which generates an explanation for each input graph, and the other is the \emph{predictor} making predictions based on the generated explanation \cite{wu2021discovering,liu2022graph}. 

Although the self-explaining methods in GNN are promising, they still suffer from heavily relying on extensive training data, which restricts their applicability in situations with limited data sizes.
For instance, during new drug discovery processes, clinical trials are conducted to assess various drug attributes such as toxicity and side effects. Due to safety concerns, the number of participants in these trials is restricted, resulting in limited experimental data. In such few-shot scenarios, existing self-explaining models fail to achieve satisfactory performance, while existing few-shot learning methods are lack of explainability. Hence, there is a pressing need to design a self-explaining GNN for few-shot scenarios.

Drawing on the fundamental human intelligence traits of rapid learning and self-explainability \cite{posner1990attention,snell2017prototypical,finn2017model}, we develop \textbf{M}eta-learned \textbf{S}elf-\textbf{E}xplaining GNN (MSE-GNN) for few-shot scenarios:

\begin{enumerate}[I.]

    \item \label{i} During classification tasks, humans initially concentrate on regions that contain crucial features, and subsequently perform classification based on these features, adhering to a two-stage paradigm \cite{posner1990attention}.
    
    \item \label{ii} When learning new concepts, humans tend to seek representative instances or prototypes and compare new instances with these prototypes to categorize them \cite{snell2017prototypical}.
    
    \item \label{iii} Humans can learn meta-knowledge from a multitude of tasks, which enables them to achieve impressive performance on new tasks with limited data, which is called \emph{``learn to learn''} \cite{finn2017model}.
    
\end{enumerate}

By incorporating these attributes into our MSE-GNN, we aim to solve the explainability of GNNs in few-shot scenarios, and then enhance the performance of both explanation and prediction tasks.

Specifically, the MSE-GNN model follows the two-stage paradigm as depicted in Figure \ref{paradigm}, which naturally mimics the human's two-stage recognition process as mentioned in~\ref{i}. Among them, the \emph{explainer}, which is composed of a GNN encoder and a MLP, predicts the probability of each node being selected as an explanation. Then, node representations encoded by another GNN encoder are separated into explanation and non-explanation based on the prediction of the \emph{explainer}. Subsequently, the \emph{predictor} mimics the decision-making process, which makes predictions based on the explanation with a MLP. 

Furthermore, the MSE-GNN model incorporates a novel mechanism that exploits task information to help with selecting explanations and making predictions. Prototype, as stated in \ref{ii}, has been proven to be effective to generate representative representations for each category \cite{vuorio2019multimodal,zhang2019interactive}. Therefore, in MSE-GNN, the concept of prototype is utilized in generating task information. The training framework of optimization-based meta-learning imitates the paradigm of ``\textit{learning to learn}'' in~\ref{iii}, where models can acquire meta-knowledge by learning from a vast array of tasks. One of the most popular and effective methods is MAML \cite{finn2017model} (Model-Agnostic Meta-Learning). Therefore, we design a new meta-training framework based on MAML to train MSE-GNN. 

We conduct extensive experiments on one synthetic dataset \cite{ying2019gnnexplainer} and three real datasets of graph classification tasks \cite{knyazev2019understanding,hu2020open}, which show excellent performance on both prediction and explanation generated.

\section{Problem Definition}
In this section, we will elaborate on the problem definition of our research. Following \cite{ma2020adaptive}, we form the few-shot graph classification problem as N-way K-shot graph classification. Given the dataset $\mathcal{G} = \{(G_1,y_1), (G_2,y_2), ...,(G_n,y_n)\}$, where $G_i$ denotes a graph with a node set $V_i$ and a edge set $E_i$. $n_i$ denotes the number of nodes in the graph. The structure feature is represented by an adjacency matrix $A_i \in \mathbb{R} ^{n_i \times n_i} $. The node attribute matrix is represented as $X_i \in \mathbb{R}^{n_i \times d}$, where $d$ is the dimension of the node attribute. 

Then, the dataset is splitted into $\{ G^{train}, y^{train}\}$ and $\{ G^{test}, y^{test}\}$ as training set and test set respectively according to label $y$. Where $y^{train} \bigcap y^{test} = \varnothing$. When training, a task $\mathcal{T}$ is sampled each time and each task contains support set $ D_{sup}^{train} = {(G_i^{train}, y_i^{train})}_{i=1}^{s}$ and query set $ D_{que}^{train} = {(G_i^{train}, y_i^{train})}_{i=1}^{q}$, where $s$ and $q$ stands for the size of support set and query set respectively. It is noteworthy that the same class space is shared in the same task. 

In each task, our goal is to optimize our model on the support set $D_{sup}$ and make predictions on the query set $D_{que}$. If a support set contains $N$ classes and $K$ data for each class, then we name the problem as N-way K-shot. When testing, we firstly finetune the learned model on support set $ D_{sup}^{test} = {(G_i^{test}, y_i^{test})}_{i=1}^{s}$ and then report the classification performance of finetuned model on $ D_{que}^{test} = {(G_i^{test}, y_i^{test})}_{i=1}^{q}$. Our goal of the few-shot graph classification problem is to develop a model that can obtain meta-knowledge across $\{ G^{train}, y^{train}\}$ and predicts labels for graphs in the query set in test stage $D_{que}^{test}$. 

In the explanation generation task, for each graph $G_i$, a node mask vector $m_i \in {[0,1]}^{n_i \times 1}$ is the explanation subgraph selected, a higher value means that the corresponding node is more important for making prediction and vice versa. Although selecting edges for explanation is a viable approach, in this paper we focus on node selection due to its computational complexity.

\section{The Proposed MSE-GNN}
\subsection{Architecture of MSE-GNN}

\begin{figure}[t]
	\centering
    \includegraphics[width=12cm]{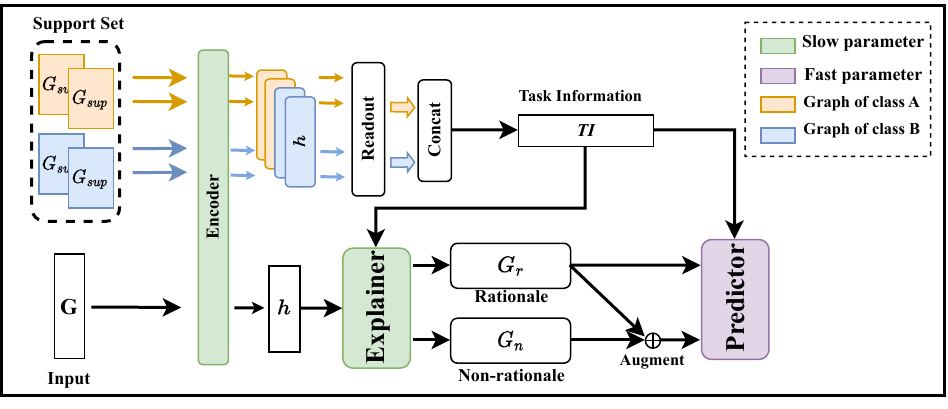}
    \caption{Overall architecture of MSE-GNN. The model employs a ``\emph{explainer-predictor}'' 2-stage self-explaining structure. The \emph{explainer} selects explanation subgraphs for each input graph. The \emph{predictor} mimics the decision-making process, which makes predictions solely based on the generated explanation.}
    \label{model}

\end{figure}

In Figure \ref{model}, we show the overall architecture of the MSE-GNN, which contains three components:  an \emph{explainer} $g$ that outputs the explanation selected, a \emph{predictor} $p$ making the final~prediction, and a graph encoder $f$. 

Before we present the details of MSE-GNN, we first clarify serveral concepts. Specifically,  existing works often combine self-explaining methods with the concept of rationale \cite{wu2021discovering,liu2022graph}. The rationale in graph data refers to the subsets of nodes or subsets of edges, which form subgraphs that determine the prediction. Hence, we posit that explanation and rationale are equivalent, as they share the same concept.

In MSE-GNN, the input graph is encoded by $f$ and each node $v$ is encoded into a node embedding $h_{(v)} \in \mathbb{R}^d$, where $d$ stands for the dimension of hidden size. The encoder can be any kind of GNN, e.g. GCN \cite{kipf2016semi}, GIN \cite{xu2018powerful}, and  GraphSAGE \cite{hamilton2017inductive}. The selector outputs a mask vector $m$ for each graph as an explanation, which divides the graph into rationale (explanation) $G_r$ and non-rationale $G_n$. Then the \emph{predictor} makes predictions based on the graph embedding rationale subgraph. Meanwhile, augmented graphs that combine rationale and non-rationale from different graphs are fed into the \emph{predictor} to ensure the robustness of the \emph{predictor}. We categorize the parameters into fast parameters and slow parameters according to the timing of updating, which will be described in detail in section \ref{section:m_t}.

\subsubsection{Task Information.} MSE-GNN generates task information for the \emph{explainer} and the \emph{predictor} to facilitate explanation generation and prediction within each task, which is composed of prototypes representing each class. 

In each task, a support set is provided, which contains data from multiple classes. We aim to extract prototypes from these data that capture the characteristics of each class in the task, in order to help with task-specific selection of explanations and the classification task. Encoded by encoder $f$, each graph is represented by a matrix containing embedding of each node:
\begin{equation}
    H_i = [...,h_{(v)},...]_{v \in V_i}^T = f(G_i) \in \mathbb{R}^{|V_i| \times d}.
\end{equation}

To obtain representation for each graph $h_i$, the readout function, e.g. mean pooling is employed, to aggregate node embeddings. By leveraging the concept of prototype learning, we further fuse the graph representations of each class with another readout function. Thus, we can obtain a prototype embedding for each class:
\begin{equation}
    TI_c = f_{readout}([...,f_{readout}(H_i),...]_{y_i = c}) \in \mathbb{R}^{d}.
\end{equation}

For an N-way K-shot classification problem, the task information is formed by concatenating prototypes of N classes. It is worth noting that, task information for each input graph of both $D_{sup}$ and $D_{que}$ is composed solely of graphs in $D_{sup}$ to prevent label leakage.

\subsubsection{Explainer.} 
The \emph{explainer} is responsible for choosing the explanation subgraph corresponding to each input graph. Specifically, given an input graph $G_i$, the \emph{explainer} firstly uses another GNN encoder to map each node to another node embedding $h_{(v)}'$ for each node in $V_i$ for selection. Then, a MLP is utilized to transform the node embeddings into a soft mask vector $m_i \in {[0,1]}^{n_i \times 1}$, with task information ${TI}_c$ and node embedding $h_{(v)}'$ concatenated as input:
\begin{equation}
    m_i = \sigma(MLP([...,[h_{(v)}',TI],...]_{v \in V_i}^T)), 
\end{equation}
where $\sigma$ denotes the sigmoid function. Hence, we can decompose the input graph $G_i$ into a rationale subgraph and non-rationale subgraph according to $m_i$ respectively:
\begin{equation}
G^r_i = \{A_i, X_i \odot m_i\} \qquad G^n_i = \{A_i, X_i \odot \overline{m_i}\},
\end{equation}
where $\overline{m_i} = \mathbf{1} - m_i$. Meanwhile, given the node embedding $h_{(v)}$ from encoder $f$, we can obtain the graph embedding for $G^r_i$ and $G^n_i$:
\begin{equation}
    h^r_i = f_{readout}(H_i \odot m_i)\qquad h^n_i = f_{readout}(H_i \odot \overline{m_i}).
\end{equation}

\subsubsection{Predictor and Graph Augmentation.}
The \emph{predictor} takes the graph embedding $h$ as input and makes the final prediction $\hat{y} = p(h)$ with a MLP. Moreover, we enhance the robustness of the \emph{predictor} through graph augmentation. Specifically, within the input graph, the rationale component represents the crucial part that determines the category, while the non-rationale component represents the noisy part. By combining the rationale and non-rationale from different graphs in the same task, additional data with noise are generated. Then we assign the label based on rationale. This approach allows us to increase the amount of noisy data, thereby improving the robustness of the \emph{predictor}. We do the combination operation by adding subgraph embeddings:
\begin{equation}
    h_{(i,j)} = h_i^r + h_j^n \qquad y_{(i,j)} = y_i,
\end{equation}
where $h_i^r$ denotes rationale from $G_i$ and $h_j^n$ means the non-rationale from $G_j$.

Therefore, in addition to task information $TI$, the \emph{predictor} $p$ receives the embeddings of both the rationale subgraphs $h_i^r$and the artificially augmented graphs $h_{(i,j)}$ for optimization, and the output are denoted as $\hat{y_i}$ and $\hat{y_{(i,j)}}$ respectively.
\begin{algorithm}[t]
	\caption{Meta-training of MSE-GNN.}
	\label{alg:algorithm}
    
    \renewcommand{\algorithmicrequire}{\textbf{Input:}}
    
    \renewcommand{\algorithmicensure}{\textbf{Output:}}
        
	\algorithmicrequire{Distribution over meta-training tasks: $p(\mathcal{T})$; Local learning rate: $\eta_{1}$; Global learning rate: $\eta_{2}$; Local update times: $T$. } \\
	\algorithmicensure{Meta-trained parameters for encoder and explanation selector:$\theta_f$, $\theta_g$, and initialization of parameters for \emph{predictor} $\theta_p$}  
    \begin{algorithmic}[1]
	\STATE Initialize $\theta = \{\theta_f, \theta_g,\theta_p\}$ randomly; 
 
    \WHILE{\textnormal{not converged}}
    
        \STATE Sample task $\mathcal{T}$ with support graphs $D_{sup}^{train}$ and query graphs $D_{que}^{train}$.
        
        \STATE Set fast adaptation parameters: $\theta_p' = \theta_p$
        
        \FOR{t  =  0 $\rightarrow$ T}
        
            \STATE Evaluate $\nabla_{\theta_p} \mathcal{L}_{sup}(\theta_f, \theta_g,\theta_p')$ by calculating loss via Equation \ref{eq:l}

            \STATE Update $\theta_p'$ : $\theta_p' \leftarrow \theta_p' - \eta_1 \cdot \nabla_{\theta_p'} \mathcal{L}_{sup}(\theta_f, \theta_g,\theta_p')$
		
        \ENDFOR
        \STATE Evaluate $\nabla_{\theta} \mathcal{L}_{que}(\theta_f, \theta_g,\theta_p')$ by calculating loss via Equation \ref{eq:l}
        
        \STATE Update $\theta$ : $\theta \leftarrow \theta - \eta_2 \cdot \nabla_{\theta} \mathcal{L}_{que}(\theta_f, \theta_g,\theta_p')$
    \ENDWHILE
\end{algorithmic}  

\end{algorithm}

\subsection{Optimization Objective}
The optimization objective of MSE-GNN is to achieve both high accuracy in predictions and generate precise explanations, which reveal the underlying reasons behind the predictions. Therefore, we design several types of loss functions and constraints. For the sake of simplicity, we consider a binary classification task without loss of generality.

With the prediction of each rationale graph embedding $p(h_i)$ and corresponding ground-truth label $y_i$, the loss function is defined as:
\begin{equation}
    L_i^r = y_i log(\hat{y_i}) + (1-y_i)log(1-\hat{y_i}).
    \label{lr}
\end{equation}

For the artificially augmented graph, our aim is to minimize the prediction values for instances of the same category while maximizing the prediction values for instances of different categories. To achieve this, we employ a contrastive loss function. For example, for a 2-way K-shot classification task, we can obtain $4K^2$ augmented graphs, where each rationale graph is combined with other $2K-1$ non-rationales, then the loss is computed as:
\begin{small}
\begin{equation}
     L_i^a = -\frac{1}{2k-1} \sum_{j=1}^{j=2K} 1_{i \neq j} \cdot 1_{y_i = y_j} \log \frac{\exp (\hat{y_i} \cdot \hat{y_j}) / \tau}{\sum_{k=1}^{k=K} 1_{i \neq k} \exp (\hat{y_i} \cdot \hat{y_j}) / \tau },
     \label{la}
\end{equation}
\end{small}
where $\tau$ is a scalar temperature hyperparameter. 

Besides, to address the deviation in the size of rationales, we introduce a penalty based on the number of rationale nodes, the following regularization term is utilized:
\begin{equation}
    L^{reg} = |\frac{1_N^\top \cdot m_i}{n_i} - \gamma|,
    \label{reg}
\end{equation}
where $\gamma$ is manually set to control the rationale size. Finally, the total loss function can be formulated as:
\begin{equation}
    L = \alpha_r \cdot L^r + \alpha_a \cdot L^a + \alpha_{reg} \cdot L^{reg},
    \label{eq:l}
\end{equation}
where $\alpha_r$, $\alpha_a$, and $\alpha_{reg}$ are hypermeters controlling the weight of each loss. 


\subsection{Meta Training}
\label{section:m_t}
Inspired by the concept of \emph{``learn to learn''} \cite{finn2017model}, we propose a new meta-training framework based on MAML to obtain meta knowledge from various tasks. We denote $\theta_f$, $\theta_g$, and $\theta_p$ as the parameters of encoder, explanation selector, and the \emph{predictor}. Specifically, MSE-GNN is trained from two procedures. One is global update, which aims to learn the parameters of encoder $\theta_f$, explanation generator $\theta_g$, and initialization of the \emph{predictor} $\theta_p$ from different tasks, the other is called local update, which performs fast adaption on new tasks and locally update only parameters of the \emph{predictor} $\theta_p'$ within each task. According to the timing of updating, we categorize the parameters into fast parameters ($\theta_p$) and slow parameters ($\theta_f$ and $\theta_g$), as shown in Figure \ref{model}.

\begin{table*}[t]
    \centering
    \renewcommand\arraystretch{0.9}
    \begin{tabular}{ccccc}
    \hline 
    & Synthetic & MNIST-sp & Molsider & Moltox21 \\ \hline
    \# Graphs & 10,000 & 70,000 & 1,427 & 7,831 \\ 
    Avg \# nodes & 74.5 & 75.0 & 33.6 & 18.6 \\ 
    Avg \# edges & 237.8 & 777.0 & 70.7 & 38.6 \\ \hline 
    \# Train tasks / classes & 5 & 5 & 19 & 7 \\ 
    \# Validate tasks / classes & 2 & 2 & 3 & 2 \\ 
    \# Test tasks / classes & 3 & 3 & 5 & 3 \\  \hline
    \end{tabular}
    \caption{Statistics of four datasets. }
    \label{tab:datasets}
   
\end{table*}

The meta-training process is demonstrated in Algorithm~\ref{alg:algorithm}. Firstly, we sample a task composed of support $D_{sup}^{train}$ and query data $D_{que}^{train}$ for each episode. Then adaption is operated by updating $\theta_p$ for T times on $D_{sup}^{train}$, where T is a hyperparameter controlling the number of local updates, which is shown in lines 5-8. With updated $\theta_p'$, we utilize the loss on $D_{que}^{train}$ to update $\theta_f$, $\theta_g$ and $\theta_p$. 

It is important to highlight that, the \emph{explainer} is trained from a variety of tasks and frozen when optimizing each task, which ensures the stability of the explanation selected across different tasks and prevents over-fitting. Therefore, $\theta_f$ and $\theta_g$ are only updated in the global update and fixed in the local update. While the \emph{predictor} needs to learn the relationship between features and categories in different tasks based on the generated explanations. As a result, the $\theta_p$ is optimized in the local update to learn the association between features and categories. Hyperparameters of loss computation in line 6 and line 9 can be differently set according to the goal of local and global optimization.

\section{Experiments}
\subsection{Datasets and Experimental Setup}
\subsubsection{Dataset.} We conduct extensive experiments on four datasets to validate the performance of MSE-GNN: (i) \textbf{Synthetic}: Due to the lack of graph datasets with explanation ground-truth, following \cite{ying2019gnnexplainer}, we create a synthetic dataset for classification, which contains 10 classes and 500 samples for each class. Each graph is composed of two parts: the rationale part and the non-rationale part. The label of each graph is determined by the rationale part. Therefore, the ground-truth of the explanation subgraph is the rationale part of each graph. (ii) \textbf{MNIST-sp} \cite{knyazev2019understanding}: MNIST-sp takes the MNIST images and transforms them into 70,000 superpixel graphs. Each graph consists of 75 nodes and is assigned one of 10 class labels. The subgraphs that represent the digits can be interpreted as ground truth explanations. (iii) \textbf{OGBG-Molsider} and \textbf{OGBG-Moltox21} \cite{hu2020open}: These two datasets are molecule datasets from the graph property prediction task on Open Graph Benchmark (OGBG), they contain 27 and 12 binary labels for each graph, which transformed into 27 and 12 binary classification tasks respectively. The dataset statistics are available in Table \ref{tab:datasets}.

\begin{table}[t]
    \centering
    \renewcommand\arraystretch{1.3}
    \resizebox{1.0\columnwidth}{!}{
    \begin{tabular}{ccccccccc}
    
    \hline
             &   \multicolumn{4}{c}{\centering Accuracy}   & \multicolumn{4}{c}{\centering AUC-ROC} \\ \hline
            &   \multicolumn{2}{c}{\centering Synthetic} & \multicolumn{2}{c}{\centering MNIST-sp} & \multicolumn{2}{c}{\centering OGBG-molsider} & \multicolumn{2}{c}{\centering OGBG-moltox21} \\ 
            & GIN &  GraphSAGE & GIN &  GraphSAGE & GIN &  GraphSAGE & GIN &  GraphSAGE \\ \hline 
          ProtoNet  & $0.8284_{\pm{0.058}}$  & $0.8327_{\pm{0.027}}$  & $0.5736_{\pm{0.008}}$  & $0.6575_{\pm{0.034}}$ & $0.5540_{\pm{0.006}}$   & $0.5468_{\pm{0.006}}$& $0.6614_{\pm{0.009}}$    & $0.6495_{\pm{0.008}}$   \\ 
          MAML      & $0.8259_{\pm{0.007}}$  & $0.6409_{\pm{0.327}}$  & $0.6283_{\pm{0.012}}$  & $0.6722_{\pm{0.009}}$ & $0.6219_{\pm{0.005}}$  & $0.6538_{\pm{0.016}}$ & $0.7217_{\pm{0.030}}$    & $0.6965_{\pm{0.014}}$   \\ 
          ASMAML    & $0.8911_{\pm{0.010}}$   & $0.7849_{\pm{0.014}}$ & $0.6526_{\pm{0.004}}$  & $0.6699_{\pm{0.023}}$ & $0.6288_{\pm{0.007}}$  & $\bm{0.6818_{\pm{0.008}}}$ & $0.7432_{\pm{0.030}}$   & $0.7181_{\pm{0.017}}$    \\ 
          \hline
          GREA\_Raw      & $0.6970_{\pm{0.005}}$  & $0.6970_{\pm{0.020}}$  & $0.6405_{\pm{0.009}}$  & $0.6667_{\pm{0.009}}$ & $0.5210_{\pm{0.009}}$  & $0.5180_{\pm{0.007}}$ & $0.5654_{\pm{0.015}}$   & $0.5479_{\pm{0.006}}$    \\
          CAL\_Raw       & $0.7248_{\pm{0.006}}$  & $0.7488_{\pm{0.007}}$  & $0.6498_{\pm{0.006}}$  & $0.6670_{\pm{0.010}}$ & $0.5978_{\pm{0.044}}$  & $0.6230_{\pm{0.008}}$ & $0.6161_{\pm{0.064}}$  & $0.6814_{\pm{0.014}}$     \\ 
          \hline
          GREA\_Meta& $0.8728_{\pm{0.013}}$  & $0.9180_{\pm{0.002}}$  & $0.6537_{\pm{0.009}}$  & $0.7430_{\pm{0.008}}$ & $0.6542_{\pm{0.005}}$  & $0.6303_{\pm{0.008}}$ & $0.7650_{\pm{0.004}}$   & $0.7582_{\pm{0.007}}$    \\ 
          CAL\_Meta & $0.8451_{\pm{0.021}}$  & $0.9096_{\pm{0.003}}$  & $\bm{0.6888_{\pm{0.007}}}$  & $\bm{0.7445_{\pm{0.019}}}$ & $0.6580_{\pm{0.012}}$  & $0.6553_{\pm{0.018}}$ & $0.7442_{\pm{0.012}}$   & $0.7652_{\pm{0.005}}$    \\  \hline
          MSE-GNN      & $\bm{0.9103_{\pm{0.004}}}$  & $\bm{0.9200_{\pm{0.004}}}$  & $0.6515_{\pm{0.008}}$  & $0.7309_{\pm{0.009}}$ & $\bm{0.6673_{\pm{0.007}}}$  & $0.6587_{\pm{0.002}}$ & $\bm{0.7735_{\pm{0.006}}}$   & $\bm{0.7728_{\pm{0.011}}}$    \\ \hline
          
    \end{tabular}
    }

    \caption{2-way 5-shot Classification Performance with a standard deviation of baseline methods and MSE-GNN.}
    \label{tab:classification}

\end{table}

\subsubsection{Experimental Setup.}
To investigate whether generating explanations can help with the classification task, we chose three few-shot learning methods: ProtoNet \cite{snell2017prototypical}, MAML \cite{finn2017model}, ASMAML \cite{ma2020adaptive}. To compare with existing self-explaining methods, we selected two state-of-the-art self-explaining models: GREA \cite{liu2022graph} and CAL \cite{sui2022causal} as baselines to compare the performance of classification and quality of generated explanations.  Moreover, for fairness, we adapt meta-training to GREA \cite{liu2022graph} and CAL \cite{sui2022causal}, enabling them to adapt to few-shot scenarios, which are denoted as GREA\_Meta and CAL\_Meta respectively. 

We use GIN and  GraphSAGE as GNN backbones for all methods. The performance of all models is evaluated on $D_{que}^{test}$. For the Synthetic and MNIST-sp with explanation ground-truth, we use Accuracy to evaluate the classification performance and AUC-ROC to evaluate the quality of the explanation selected. For the two molecule datasets, due to the absence of explanation ground-truth, we only evaluate the classification performance using Area under the ROC curve (AUC) following \cite{liu2022graph}. For meta-training, we utilize Adam optimizer for local and global updates and set local update times $T$ to 5. Local learning rate $\eta_1$ is set to 0.001 and global learning rate $\eta_1$ is tuned over \{1e-5, 1e-4, 1e-3\}. $\gamma$ in Equation \ref{reg} is tuned over \{0.1, 0.2, 0.3, 0.4, 0.5\}, number of GNN layers is tuned over \{2,3\}. We select hyper-parameters based on related works and grid searches. All our experiments are conducted with one Tesla V100 GPU.

\subsubsection{Performance on Synthetic Graphs and MNIST-sp.}
To explore whether MSE-GNN can achieve high performance on classification and generate high-quality explanation, we conduct 2-way 5-shot experiments on Synthetic and MNIST-sp datasets which contain ground-truth explanations for each graph. The experimental results are summarized in Table \ref{tab:classification} and Table \ref{tab:explanation}. We first compare meta-trained self-explaining baseline models (GREA\_Meta, CAL\_Meta) with themselves (GREA\_Raw, CAL\_Raw). We can observe that significant performance boosts are brought by meta-training on both classification and explanation, which indicates that meta-training can leverage the meta-knowledge learned across training tasks effectively on new tasks.

\begin{table}[t]
    \centering
    \renewcommand\arraystretch{1.1}
    \scalebox{0.73}{
    \begin{tabular}{cccc}
 
          \hline
        
               &       & Synthetic & MNIST-sp  \\ \hline

       \multirow{5}{*}{GIN}   & GREA\_Raw      & $0.4934_{\pm{0.006}}$   & $0.4789_{\pm{0.044}}$ \\ 
         & CAL\_Raw       & $0.4741_{\pm{0.0250}}$   & $0.4395_{\pm{0.039}}$        \\ 
          & GREA\_Meta& $0.6745_{\pm{0.0265}}$   & $0.7855_{\pm{0.013}}$        \\ 
          & CAL\_Meta & $0.6201_{\pm{0.0550}}$   & $0.1707_{\pm{0.0243}}$        \\ 
          & MSE-GNN      & $\bm{0.7000_{\pm{0.006}}}$   & $\bm{0.8222_{\pm{0.030}}}$        \\ \hline
       \multirow{5}{*}{Graghsage}   & GREA\_Raw      & $0.4929_{\pm{0.023}}$   & $0.5496_{\pm{0.064}}$  \\ 
         & CAL\_Raw       & $0.5080_{\pm{0.054}}$    & $0.4906_{\pm{0.116}}$         \\ 
          & GREA\_Meta& $0.7099_{\pm{0.014}}$    & $0.6513_{\pm{0.040}}$         \\ 
          & CAL\_Meta & $0.6858_{\pm{0.015}}$    & $0.6613_{\pm{0.229}}$        \\ 
          & MSE-GNN      & $\bm{0.7189_{\pm{0.012}}}$   & $\bm{0.7077_{\pm{0.038}}}$        \\ \hline

    \end{tabular}
    }

    \caption{For the Synthetic and MNIST-sp with explanation ground-truth, AUC-ROC is utilized to evaluate the quality of the explanation selected.}
 
    \label{tab:explanation}

\end{table}

On Synthetic, MSE-GNN shows superiority to other baseline methods on the performance of classification and explanation quality. Compared to meta-trained self-explaining baselines, MSE-GNN performs better on both classification and explanation as MSE-GNN utilizes task information and effectively leverages the augmented graph through the introduction of supervised contrastive loss. Moreover, the inherent denoising capability of self-explaining models contributes to the superior classification performance of MSE-GNN compared to ProtoNet, MAML, and ASMAML.

\begin{figure}[t]
	\centering
    \includegraphics[width=8.7cm]{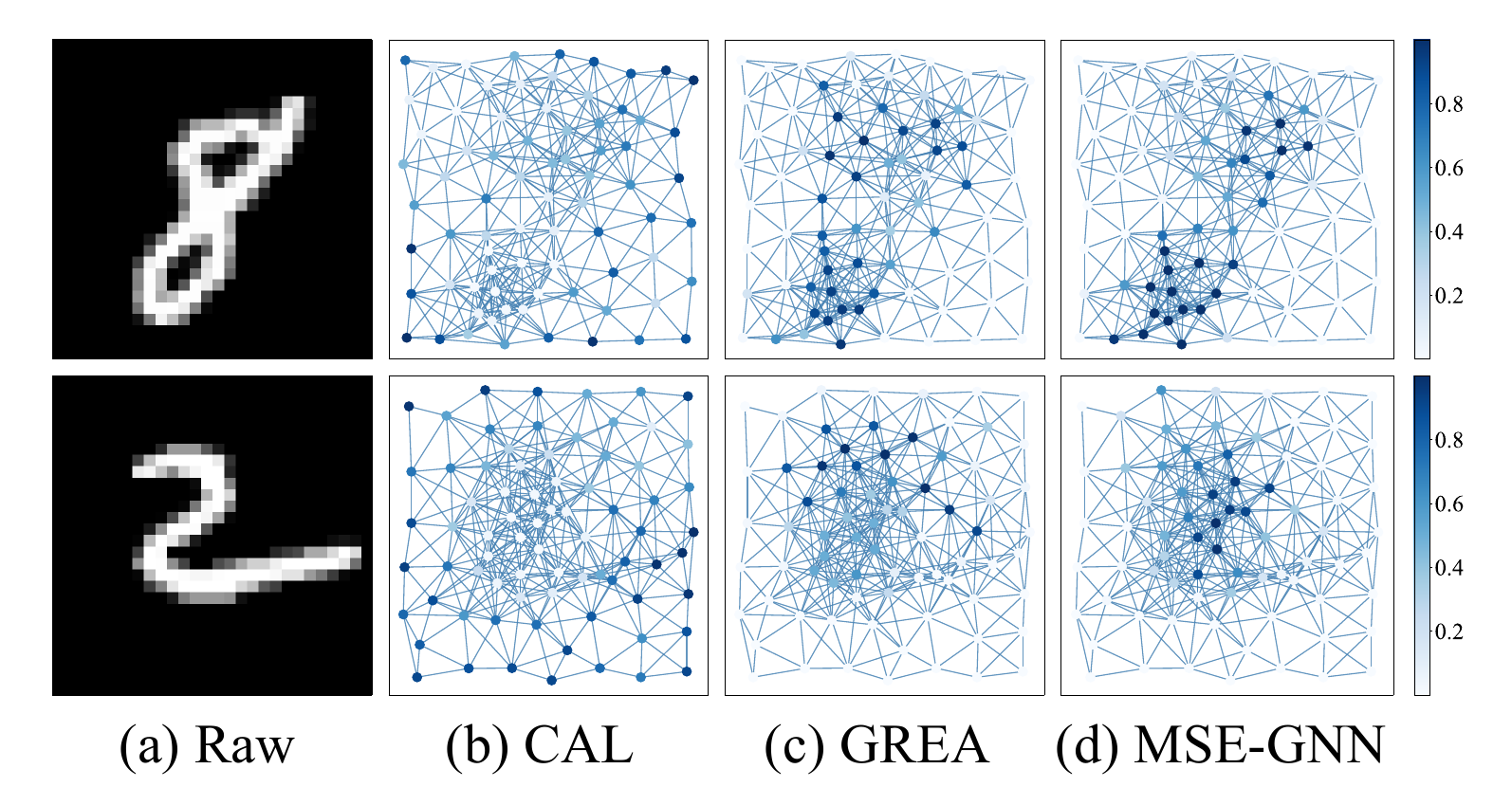}
    \caption{Raw figure of MNIST-sp and visualization of explanations generated by CAL(a), GREA(b) and MSE-GNN(c). Darker nodes indicate higher importance scores.}
    \label{fig:vis_mnist}

\end{figure}

Unexpectedly, CAL achieves the best classification performance on MNIST-sp, especially when using GIN as the backbone, surpassing MSE-GNN by over 5\%. Meanwhile, the quality of explanations is significantly lower compared to GREA and MSE-GNN. By visualization in Figure \ref{fig:vis_mnist}, which reveals the internal reasoning process of models, we can find that CAL generated explanations that were opposite to our expectations, indicating that CAL infers the digit based on the shape of the background. It is also easy to understand that the digital in a picture can be inferred from the background since the number part and the background part are complementary sets. Therefore, despite the generated explanations being contrary to our expectations, CAL's performance demonstrated that utilizing background information for digit prediction is more effective on MNIST-sp. The reason for CAL generating opposite explanations is that it lacks constraints on the size of the explanation. As a result, it tends to favor subgraphs that contain more useful information and overlook the size of the explanation subgraph. Furtherly comparing the visualization of explanations of MSE-GNN and GREA, we can find that explanations of MSE-GNN are more compact and focus more on the digital part, which is in line with the result in Table \ref{tab:explanation}.

\begin{figure}[t]
	\centering
	\begin{subfigure}{0.32\linewidth}
		\centering
		\includegraphics[width=0.99\linewidth]{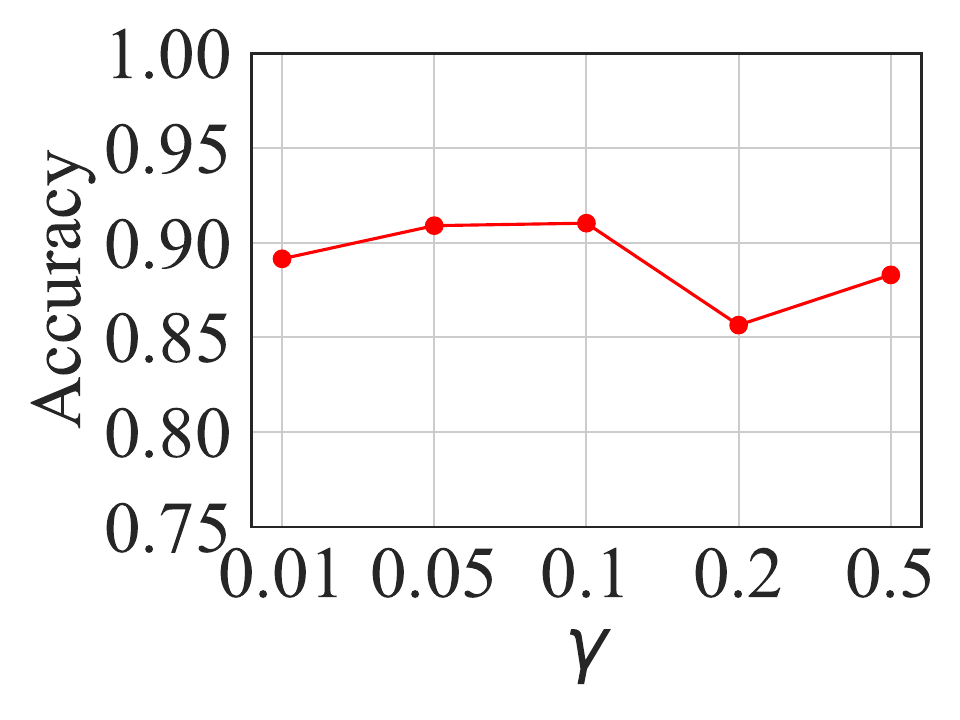}
 
		\caption{Classification Performance on Synthetic}
		\label{gamma1}
	\end{subfigure}
	\centering
	\begin{subfigure}{0.32\linewidth}
		\centering
		\includegraphics[width=0.99\linewidth]{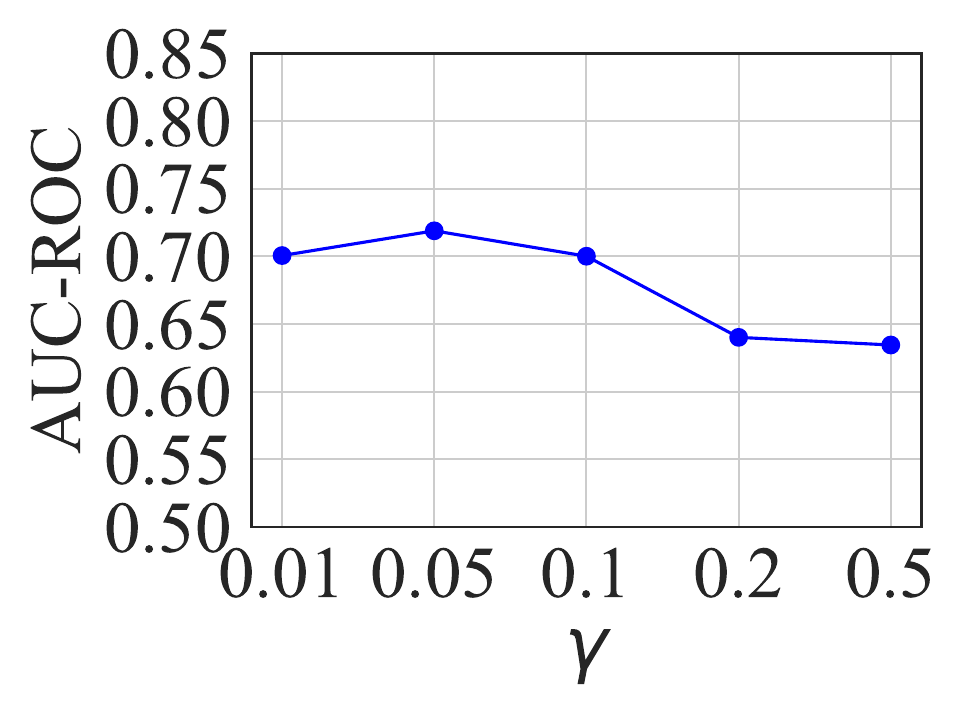}

		\caption{Quality of explanation on Synthetic}
		\label{gamma2}
	\end{subfigure}
	\centering
	\begin{subfigure}{0.32\linewidth}
		\centering
		\includegraphics[width=0.99\linewidth]{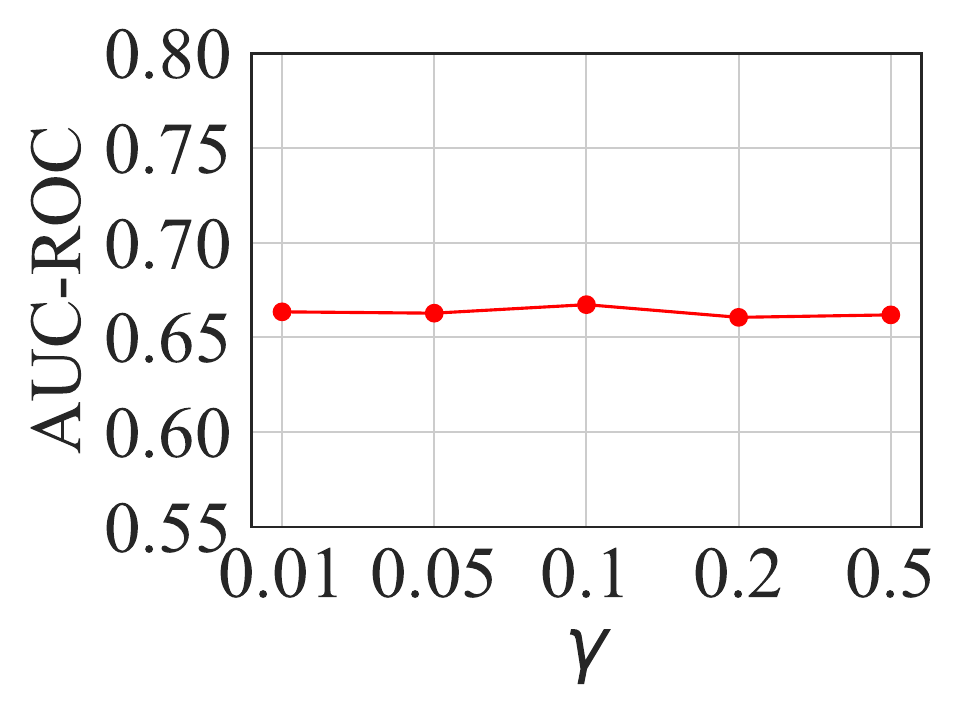}

		\caption{Classification Performance on OGBG-Molsider}
		\label{gamma3}
	\end{subfigure}
	\caption{Classification Performance and quality of explanation selected on Synthetic and OGBG-Molsider with different $\gamma$.}

	\label{fig: dif_gamma}

\end{figure}

\subsubsection{Performance on OGBG.}
MSE-GNN achieves comparable classification performance on these two molecule datasets, demonstrating the effectiveness of its structure. Furthermore, we can observe that the self-explaining models with meta-training outperform all meta-learning models except on OGBG-molsider using  GraphSAGE. This is because the process of generating explanations can potentially improve the classification task by eliminating irrelevant noise.


\subsubsection{Performance with Different Size of Support Set.}
\begin{figure}[htbp]
	\centering

    \includegraphics[width=0.95\linewidth]{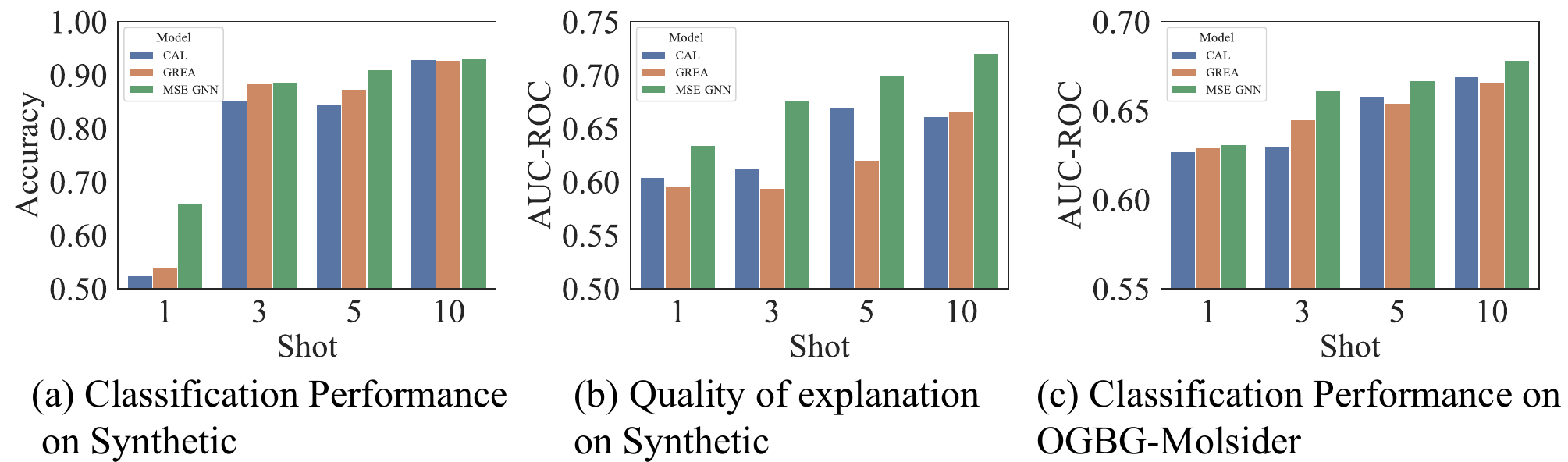}

    \caption{Classification Performance and quality of explanation selected on Synthetic and OGBG-Molsider with different size of support sets.}
    \label{fig: dif_shot}

\end{figure}

Intuitively, for a classification task, the size of the training set has a significant impact on the model's performance. Therefore, in the scenario of few-shot learning, we evaluate the performance of MSE-GNN and other self-explaining models under different support set sizes. Experimental results are shown in Figure \ref{fig: dif_shot}. First, comparing different methods, we observe that MSE-GNN consistently outperforms other baselines across different support set sizes, which further validates the performance of MSE-GNN on both classification and explaining. Next, comparing the performance of MSE-GNN across different support set sizes, we observe that as the support set size increases, both the classification accuracy and the quality of generated explanations improve. This also demonstrates the importance of training set size on model performance.

\subsubsection{Ablation Study.}
Table \ref{tab:ablation} demonstrates the impact of contrastive loss and task information utilized in MSE-GNN on Synthetic with GIN. When applying Contrastive Loss (CL), both the classification accuracy and the quality of generated explanations of the model are improved. This indicates that introducing contrastive loss can enhance the model's performance and lead to better results in prediction and explanation tasks. On the other hand, when applying Task Information (TI), the model's performance is also improved across all datasets. This suggests that incorporating task information into the model can provide additional context and guidance, thereby enhancing the model's ability. Moreover, when both CL and TI are used together, the model excels significantly across all datasets, indicating that the combination of CL and TI can synergistically contribute to better performance on both classification and explanation tasks.


\subsubsection{Sensitivity Analysis.}
In MSE-GNN, the parameter $\gamma$ is crucial in controlling the size of the selected explanation. To examine the sensitivity of the model to different values of $\gamma$, we conduct a sensitivity analysis on the Synthetic and OGBG-Molsider datasets with GIN. As illustrated in Figure \ref{fig: dif_gamma}, the results demonstrate that MSE-GNN achieves the best classification performance when $\gamma$ is set to 0.1 on both datasets, while the explaining performance achieves best when $\gamma$ equals 0.05 on Synthetic. We observe that as the value of $\gamma$ deviates from these two optimal points, the classification performance or the quality of generated explanations decreases. We also notice that the impact of $\gamma$ is less pronounced on the OGBG-Molsider dataset, indicating that the model is less sensitive to $\gamma$ on OGBG-Molsider.

\begin{figure}[t]
	\centering
	\begin{subfigure}{0.32\linewidth}
		\centering
		\includegraphics[width=0.99\linewidth]{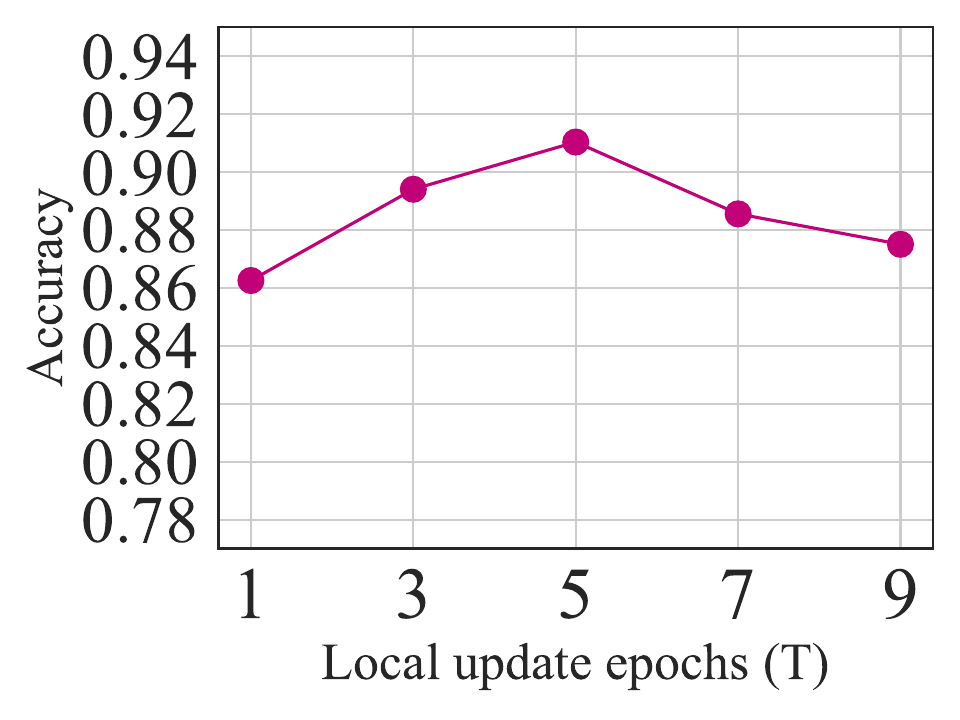}
 
		\caption{Classification Performance on Synthetic}
		\label{gamma1}
	\end{subfigure}
	\centering
	\begin{subfigure}{0.32\linewidth}
		\centering
		\includegraphics[width=0.99\linewidth]{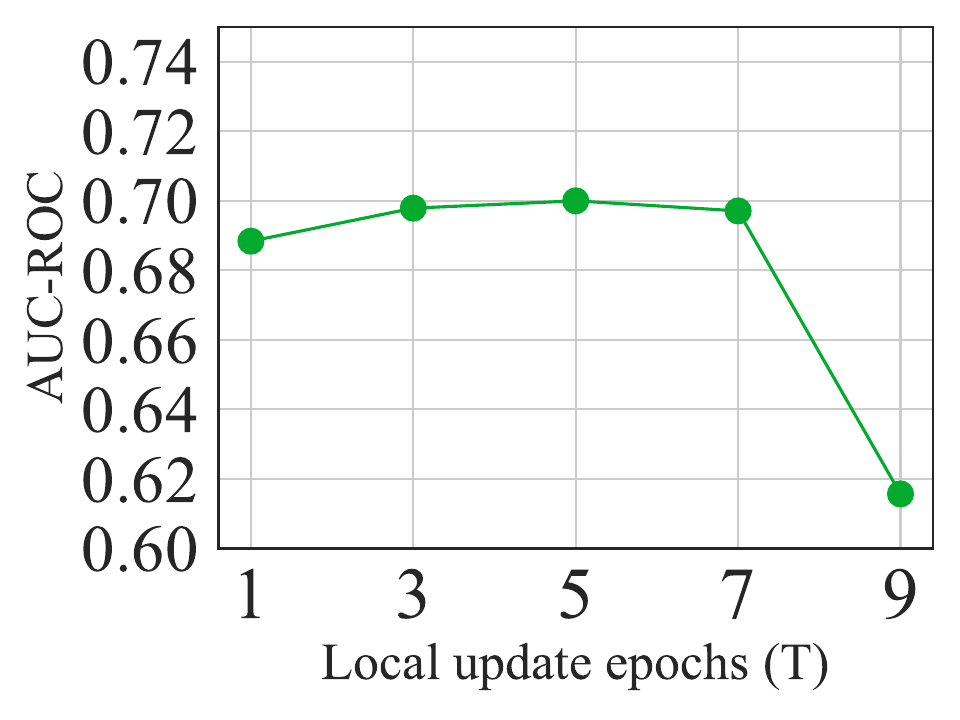}

		\caption{Quality of explanation on Synthetic}
		\label{gamma2}
	\end{subfigure}
	\centering
	\begin{subfigure}{0.32\linewidth}
		\centering
		\includegraphics[width=0.99\linewidth]{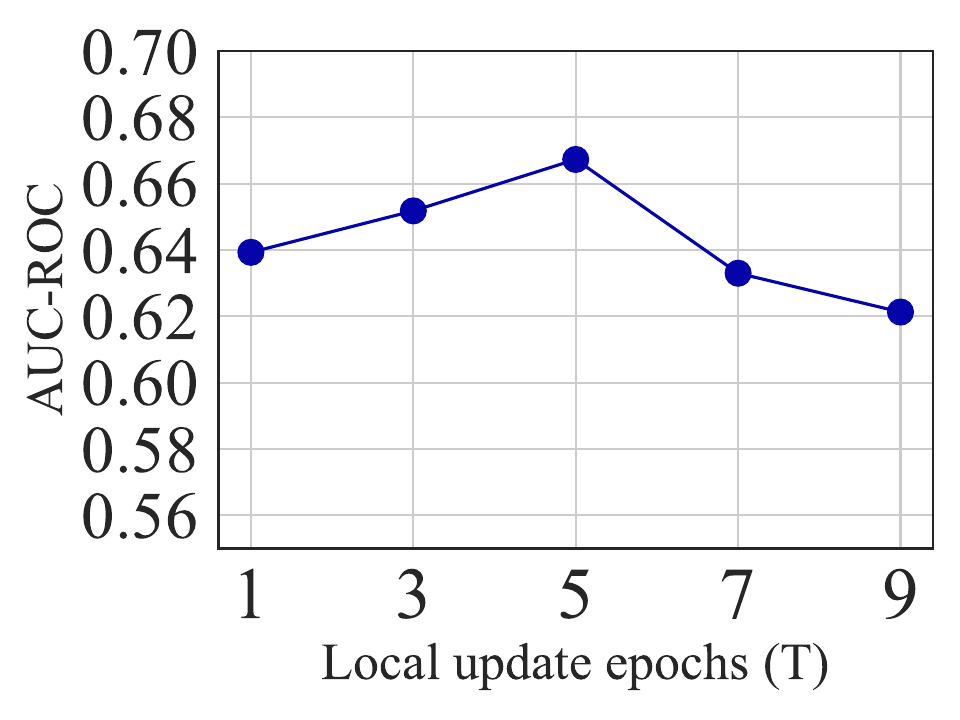}

		\caption{Classification Performance on OGBG-Molsider}
		\label{gamma3}
	\end{subfigure}
	\caption{Classification Performance and quality of explanation selected on Synthetic and OGBG-Molsider with different $T$.}

	\label{fig: dif_epochs}

\end{figure}

Furthermore, $T$, which stands for the number of local update epochs, affects both the effectiveness and efficiency of the MSE-GNN. We compared the performance of MSE-GNN with different local update epochs on the Synthetic and OGBG-Molsider datasets. The experimental results shown in Figure \ref{fig: dif_epochs} indicate that when $T$ is set to 5, MSE-GNN achieves the best classification and explaining performance on both Synthetic and OGBG-molsider. A too-small (too-large) $T$ may result in underfitting (overfitting) of the model for new tasks.

\begin{table}[t]
    \centering
    \renewcommand\arraystretch{1.2}
    \scalebox{0.8}{
    \begin{tabular}{ccccc}

    \hline
    \multirow{2}{*}{\centering CL} & \multirow{2}{*}{\centering TI} & \multicolumn{2}{c}{Synthetic} &  OGBG-molsider \\ 
      
         &  &Classif. & Explan. & Classif. \\ \hline
         &  & $0.8728_{\pm{0.013}}$ & $0.6745_{\pm{0.027}}$  & $0.6542_{\pm{0.005}}$ \\ 
         \checkmark & &$0.8809_{\pm{0.037}}$ & $0.6860_{\pm{0.028}}$ & $0.6623_{\pm{0.011}}$\\ 
         &\checkmark  & $0.8800_{\pm{0.011}}$& $0.6766_{\pm{0.014}}$& $0.6616_{\pm{0.001}}$\\ 
         \checkmark&\checkmark  & $\bm{0.9103_{\pm{0.004}}}$ &$\bm{0.7000_{\pm{0.006}}}$ & $\bm{0.6673_{\pm{0.007}}}$      \\ \hline

    \end{tabular}
    }

    \caption{Impact of contrastive loss and task information.}
    \label{tab:ablation}

\end{table}

\section{Related Works}
\subsubsection{Few-shot learning and Meta Learning on Graph Classification}
Few-shot learning aims to learn a model with only a few samples. A promising kind of method is meta learning. Meta learning is also known as ``learning to learn", which attempts to learn meta-knowledge from a variety of tasks. There two catogeries for meta-learning \cite{zhang2022few}: metric-based models \cite{snell2017prototypical,chauhan2019few,gao2023leveraging,niu2021relational,wang2021reform} and optimization-based models \cite{finn2017model,guo2021few,zhou2019meta,ma2020adaptive,wang2021property}. The former focuses on computing the distance between query data and class prototypes \cite{snell2017prototypical}. The latter aims to learn an effective initialization of parameters, which enables rapid adaption \cite{finn2017model}. \cite{zhou2019meta} firstly applied meta learning framework to the node classification task. \cite{ma2020adaptive} utilize a step controller for the robustness and generalization of meta-learner. Notwithstanding the remarkable accuracy improvement achieved by these methods on few-shot learning tasks, their lack of explainability hinders their applicability in certain scenarios such as the medical and finance area.

\subsubsection{Explainability in Graph Neural Network}
With more attention paid to the applications of GNNs, the explainability of GNNs is more crucial. The explanation increases the models’ transparency and enhances practitioners’ trust in GNN models by enriching their understanding of why the decision is made by GNNs. Explainability of GNNs can be categorized into two classes \cite{yuan2022explainability,yue2024cooperative}: post-hoc explanations and self-explainable GNNs. Post-hoc explanations attempt to give explanations for trained GNNs with additional \emph{explainer} model ~\cite{ying2019gnnexplainer,wang2022gnninterpreter,huang2022graphlime,lucic2022cf,azzolin2022global,luo2020parameterized,duval2021graphsvx,kamal2022improving}. However, these post-hoc explainers often fail to unveil the true reasoning process of the model due to the non-convexity and complexity of the underlying GNN models \cite{rudin2018please}. Self-explaining GNNs design specific GNN models which are interpretable intrinsically ~\cite{wu2021discovering,sui2022causal,liu2022graph,zhang2022protgnn,muller2022dt+,azzolin2022global}. They output the prediction and corresponding explanation simultaneously. DIR \cite{wu2021discovering} aims to extract causal rationales that remain consistent across various distributions while eliminating unstable spurious patterns. 
GREA \cite{liu2022graph} is another self-explainable model that introduces a new augmentation operation called environment replacement that automatically creates virtual data examples to improve rationale identification. Another category of self-explaining models leverages the concept of prototype learning \cite{zhang2022protgnn,shin2022prototype,azzolin2022global,seo2024interpretable,zhang2022incorporating}. ProtGNN \cite{zhang2022protgnn} provides explanations by selecting subgraphs that are the most relevant to graph patterns for identifying graphs of each class. However, existing self-explainable GNNs overlook the scarcity of labeled graph data in many applications. Thus, it's important to build few-shot learning models with self-explainability.

\section{Conclusion}

In this paper, we proposed MSE-GNN to address the explainability of GNN in few-shot scenarios. To be specific, MSE-GNN adopted a ``\emph{explainer-predictor}'' 2-stage self-explaining structure and a meta-training framework based on meta-learning, which improved performance in few-shot scenarios. MSE-GNN also introduced a mechanism to leverage task information to assist explanation generation and result prediction. Additionally, MSE-GNN employed graph augmentation to enhance model robustness. Extensive experimental results demonstrated that MSE-GNN achieves strong performance in classification tasks while selecting high-quality explanations in few-shot scenarios.

\subsubsection{Acknowledgements.} 
This research was partially supported by supported by Anhui Provincial Natural Science Foundation (No. 2308085QF229), the Fundamental Research Funds for the Central Universities (No. WK2150110034), Technology Innovation Community in Yangtze River Delta (No. 2023CSJZN0200), and the Fundamental Research Funds for the Central Universities.

%
%
%
\bibliographystyle{splncs04}

\bibliography{ref}

\begin{thebibliography}{10}
\providecommand{\url}[1]{\texttt{#1}}
\providecommand{\urlprefix}{URL }
\providecommand{\doi}[1]{https://doi.org/#1}

\bibitem{azzolin2022global}
Azzolin, S., Longa, A., Barbiero, P., Lio, P., Passerini, A.: Global explainability of gnns via logic combination of learned concepts. In: The Eleventh International Conference on Learning Representations (2022)

\bibitem{bian2020rumor}
Bian, T., Xiao, X., Xu, T., Zhao, P., Huang, W., Rong, Y., Huang, J.: Rumor detection on social media with bi-directional graph convolutional networks. In: Proceedings of the AAAI conference on artificial intelligence. vol.~34, pp. 549--556 (2020)

\bibitem{chauhan2019few}
Chauhan, J., Nathani, D., Kaul, M.: Few-shot learning on graphs via super-classes based on graph spectral measures. In: International Conference on Learning Representations (2019)

\bibitem{chen2020revisiting}
Chen, L., Wu, L., Hong, R., Zhang, K., Wang, M.: Revisiting graph based collaborative filtering: A linear residual graph convolutional network approach. In: Proceedings of the AAAI conference on artificial intelligence. vol.~34, pp. 27--34 (2020)

\bibitem{duval2021graphsvx}
Duval, A., Malliaros, F.D.: Graphsvx: Shapley value explanations for graph neural networks. In: Machine Learning and Knowledge Discovery in Databases. Research Track: European Conference, ECML PKDD 2021, Bilbao, Spain, September 13--17, 2021, Proceedings, Part II 21. pp. 302--318. Springer (2021)

\bibitem{dwivedi2023benchmarking}
Dwivedi, V.P., Joshi, C.K., Luu, A.T., Laurent, T., Bengio, Y., Bresson, X.: Benchmarking graph neural networks. Journal of Machine Learning Research  \textbf{24},  1--48 (2023)

\bibitem{finn2017model}
Finn, C., Abbeel, P., Levine, S.: Model-agnostic meta-learning for fast adaptation of deep networks. In: International conference on machine learning. pp. 1126--1135. PMLR (2017)

\bibitem{gao2023leveraging}
Gao, W., Wang, H., Liu, Q., Wang, F., Lin, X., Yue, L., Zhang, Z., Lv, R., Wang, S.: Leveraging transferable knowledge concept graph embedding for cold-start cognitive diagnosis. In: Proceedings of the 46th International ACM SIGIR Conference on Research and Development in Information Retrieval. pp. 983--992 (2023)

\bibitem{guo2021few}
Guo, Z., Zhang, C., Yu, W., Herr, J., Wiest, O., Jiang, M., Chawla, N.V.: Few-shot graph learning for molecular property prediction. In: Proceedings of the Web Conference 2021. pp. 2559--2567 (2021)

\bibitem{hamilton2017inductive}
Hamilton, W., Ying, Z., Leskovec, J.: Inductive representation learning on large graphs. Advances in neural information processing systems  \textbf{30} (2017)

\bibitem{hu2020open}
Hu, W., Fey, M., Zitnik, M., Dong, Y., Ren, H., Liu, B., Catasta, M., Leskovec, J.: Open graph benchmark: Datasets for machine learning on graphs. Advances in neural information processing systems  \textbf{33},  22118--22133 (2020)

\bibitem{huang2022graphlime}
Huang, Q., Yamada, M., Tian, Y., Singh, D., Chang, Y.: Graphlime: Local interpretable model explanations for graph neural networks. IEEE Transactions on Knowledge and Data Engineering  (2022)

\bibitem{kamal2022improving}
Kamal, A., Vincent, E., Plantevit, M., Robardet, C.: Improving the quality of rule-based gnn explanations. In: Joint European Conference on Machine Learning and Knowledge Discovery in Databases. pp. 467--482. Springer (2022)

\bibitem{kipf2016semi}
Kipf, T.N., Welling, M.: Semi-supervised classification with graph convolutional networks. In: International Conference on Learning Representations (2016)

\bibitem{knyazev2019understanding}
Knyazev, B., Taylor, G.W., Amer, M.: Understanding attention and generalization in graph neural networks. Advances in neural information processing systems  \textbf{32} (2019)

\bibitem{lin2022orphicx}
Lin, W., Lan, H., Wang, H., Li, B.: Orphicx: A causality-inspired latent variable model for interpreting graph neural networks. In: Proceedings of the IEEE/CVF Conference on Computer Vision and Pattern Recognition. pp. 13729--13738 (2022)

\bibitem{liu2022graph}
Liu, G., Zhao, T., Xu, J., Luo, T., Jiang, M.: Graph rationalization with environment-based augmentations. In: Proceedings of the 28th ACM SIGKDD Conference on Knowledge Discovery and Data Mining. pp. 1069--1078 (2022)

\bibitem{lucic2022cf}
Lucic, A., Ter~Hoeve, M.A., Tolomei, G., De~Rijke, M., Silvestri, F.: Cf-gnnexplainer: Counterfactual explanations for graph neural networks. In: International Conference on Artificial Intelligence and Statistics. pp. 4499--4511. PMLR (2022)

\bibitem{luo2020parameterized}
Luo, D., Cheng, W., Xu, D., Yu, W., Zong, B., Chen, H., Zhang, X.: Parameterized explainer for graph neural network. Advances in neural information processing systems  \textbf{33},  19620--19631 (2020)

\bibitem{ma2020adaptive}
Ma, N., Bu, J., Yang, J., Zhang, Z., Yao, C., Yu, Z., Zhou, S., Yan, X.: Adaptive-step graph meta-learner for few-shot graph classification. In: Proceedings of the 29th ACM International Conference on Information \& Knowledge Management. pp. 1055--1064 (2020)

\bibitem{muller2022dt+}
M{\"u}ller, P., Faber, L., Martinkus, K., Wattenhofer, R.: Dt+ gnn: A fully explainable graph neural network using decision trees. arXiv preprint arXiv:2205.13234  (2022)

\bibitem{niu2021relational}
Niu, G., Li, Y., Tang, C., Geng, R., Dai, J., Liu, Q., Wang, H., Sun, J., Huang, F., Si, L.: Relational learning with gated and attentive neighbor aggregator for few-shot knowledge graph completion. In: Proceedings of the 44th International ACM SIGIR Conference on Research and Development in Information Retrieval. pp. 213--222 (2021)

\bibitem{posner1990attention}
Posner, M.I., Petersen, S.E.: The attention system of the human brain. Annual review of neuroscience  \textbf{13},  25--42 (1990)

\bibitem{pourhabibi2020fraud}
Pourhabibi, T., Ong, K.L., Kam, B.H., Boo, Y.L.: Fraud detection: A systematic literature review of graph-based anomaly detection approaches. Decision Support Systems  \textbf{133},  113303 (2020)

\bibitem{rudin2018please}
Rudin, C.: Please stop explaining black box models for high stakes decisions. stat  \textbf{1050}, ~26 (2018)

\bibitem{seo2024interpretable}
Seo, S., Kim, S., Park, C.: Interpretable prototype-based graph information bottleneck. Advances in Neural Information Processing Systems  \textbf{36} (2024)

\bibitem{shin2022prototype}
Shin, Y.M., Kim, S.W., Yoon, E.B., Shin, W.Y.: Prototype-based explanations for graph neural networks (student abstract). In: Proceedings of the AAAI Conference on Artificial Intelligence. vol.~36, pp. 13047--13048 (2022)

\bibitem{smuha2019eu}
Smuha, N.A.: The eu approach to ethics guidelines for trustworthy artificial intelligence. Computer Law Review International  \textbf{20},  97--106 (2019)

\bibitem{snell2017prototypical}
Snell, J., Swersky, K., Zemel, R.: Prototypical networks for few-shot learning. Advances in neural information processing systems  \textbf{30} (2017)

\bibitem{sui2022causal}
Sui, Y., Wang, X., Wu, J., Lin, M., He, X., Chua, T.S.: Causal attention for interpretable and generalizable graph classification. In: Proceedings of the 28th ACM SIGKDD Conference on Knowledge Discovery and Data Mining. pp. 1696--1705 (2022)

\bibitem{vuorio2019multimodal}
Vuorio, R., Sun, S.H., Hu, H., Lim, J.J.: Multimodal model-agnostic meta-learning via task-aware modulation. Advances in neural information processing systems  \textbf{32} (2019)

\bibitem{wang2021reform}
Wang, S., Huang, X., Chen, C., Wu, L., Li, J.: Reform: Error-aware few-shot knowledge graph completion. In: Proceedings of the 30th ACM International Conference on Information \& Knowledge Management. pp. 1979--1988 (2021)

\bibitem{wang2022gnninterpreter}
Wang, X., Shen, H.W.: Gnninterpreter: A probabilistic generative model-level explanation for graph neural networks. In: The Eleventh International Conference on Learning Representations (2022)

\bibitem{wang2021property}
Wang, Y., Abuduweili, A., Yao, Q., Dou, D.: Property-aware relation networks for few-shot molecular property prediction. Advances in Neural Information Processing Systems  \textbf{34},  17441--17454 (2021)

\bibitem{wieder2020compact}
Wieder, O., Kohlbacher, S., Kuenemann, M., Garon, A., Ducrot, P., Seidel, T., Langer, T.: A compact review of molecular property prediction with graph neural networks. Drug Discovery Today: Technologies  \textbf{37},  1--12 (2020)

\bibitem{wu2022graph}
Wu, L., Cui, P., Pei, J., Zhao, L., Guo, X.: Graph neural networks: foundation, frontiers and applications. In: Proceedings of the 28th ACM SIGKDD Conference on Knowledge Discovery and Data Mining. pp. 4840--4841 (2022)

\bibitem{wu2021discovering}
Wu, Y., Wang, X., Zhang, A., He, X., Chua, T.S.: Discovering invariant rationales for graph neural networks. In: International Conference on Learning Representations (2021)

\bibitem{xu2018powerful}
Xu, K., Hu, W., Leskovec, J., Jegelka, S.: How powerful are graph neural networks? In: International Conference on Learning Representations (2018)

\bibitem{ying2019gnnexplainer}
Ying, Z., Bourgeois, D., You, J., Zitnik, M., Leskovec, J.: Gnnexplainer: Generating explanations for graph neural networks. Advances in neural information processing systems  \textbf{32} (2019)

\bibitem{yuan2022explainability}
Yuan, H., Yu, H., Gui, S., Ji, S.: Explainability in graph neural networks: A taxonomic survey. IEEE Transactions on Pattern Analysis and Machine Intelligence  (2022)

\bibitem{yue2022dare}
Yue, L., Liu, Q., Du, Y., An, Y., Wang, L., Chen, E.: Dare: disentanglement-augmented rationale extraction. Advances in Neural Information Processing Systems  \textbf{35},  26603--26617 (2022)

\bibitem{yue2024cooperative}
Yue, L., Liu, Q., Liu, Y., Gao, W., Yao, F., Li, W.: Cooperative classification and rationalization for graph generalization. In: Proceedings of the ACM Web Conference. vol.~2024 (2024)

\bibitem{yue2023interventional}
Yue, L., Liu, Q., Wang, L., An, Y., Du, Y., Huang, Z.: Interventional rationalization. In: Proceedings of the 2023 Conference on Empirical Methods in Natural Language Processing. pp. 11404--11418 (2023)

\bibitem{zhang2022few}
Zhang, C., Ding, K., Li, J., Zhang, X., Ye, Y., Chawla, N.V., Liu, H.: Few-shot learning on graphs: A survey. In: The 31st International Joint Conference on Artificial Intelligence (IJCAI) (2022)

\bibitem{zhang2021eatn}
Zhang, K., Liu, Q., Qian, H., Xiang, B., Cui, Q., Zhou, J., Chen, E.: Eatn: An efficient adaptive transfer network for aspect-level sentiment analysis. IEEE Transactions on Knowledge and Data Engineering  \textbf{35}(1),  377--389 (2021)

\bibitem{zhang2019interactive}
Zhang, K., Zhang, H., Liu, Q., Zhao, H., Zhu, H., Chen, E.: Interactive attention transfer network for cross-domain sentiment classification. In: Proceedings of the AAAI Conference on Artificial Intelligence. vol.~33, pp. 5773--5780 (2019)

\bibitem{zhang2022incorporating}
Zhang, K., Zhang, K., Zhang, M., Zhao, H., Liu, Q., Wu, W., Chen, E.: Incorporating dynamic semantics into pre-trained language model for aspect-based sentiment analysis. arXiv preprint arXiv:2203.16369  (2022)

\bibitem{zhang2024fedgt}
Zhang, Z., Hu, Q., Yu, Y., Gao, W., Liu, Q.: Fedgt: Federated node classification with scalable graph transformer. arXiv preprint arXiv:2401.15203  (2024)

\bibitem{zhang2022hierarchical}
Zhang, Z., Liu, Q., Hu, Q., Lee, C.K.: Hierarchical graph transformer with adaptive node sampling. Advances in Neural Information Processing Systems  \textbf{35},  21171--21183 (2022)

\bibitem{zhang2022protgnn}
Zhang, Z., Liu, Q., Wang, H., Lu, C., Lee, C.: Protgnn: Towards self-explaining graph neural networks. In: Proceedings of the AAAI Conference on Artificial Intelligence. vol.~36, pp. 9127--9135 (2022)

\bibitem{zhou2019meta}
Zhou, F., Cao, C., Zhang, K., Trajcevski, G., Zhong, T., Geng, J.: Meta-gnn: On few-shot node classification in graph meta-learning. In: Proceedings of the 28th ACM International Conference on Information and Knowledge Management. pp. 2357--2360 (2019)

\end{thebibliography}

\end{document}